\documentclass[11pt]{article}

\usepackage[final]{acl}

\usepackage{times}
\usepackage{latexsym}

\usepackage[T1]{fontenc}
\usepackage[utf8]{inputenc}

\usepackage{microtype}

\usepackage{inconsolata}

\usepackage{graphicx}
\usepackage[whole]{bxcjkjatype}
\usepackage{colortbl,xcolor}
\usepackage{booktabs,tabularx,array}
\usepackage{multirow}
\usepackage{amsmath}

\title{I Understand How You Feel: Enhancing Deeper Emotional Support
Through Multilingual Emotional Validation in Dialogue System}

\author{
 \textbf{Zi Haur Pang},
 \textbf{Yahui Fu},
 \textbf{Koji Inoue},
 \textbf{and}
 \textbf{Tatsuya Kawahara}
\\
  Graduate School of Informatics, Kyoto University, Japan 
  \\
  \texttt{\{pang, fu, inoue, kawahara\}@sap.ist.i.kyoto-u-ac.jp}}

\begin{document}
\maketitle
\begin{abstract}
\emph{Emotional validation}  -  explicitly acknowledging that a user's feelings make sense - has proven therapeutic value but has received little computational attention.
% We introduce the first three-stage framework for emotional validation in dialogue systems, decomposing the problem into (i) validating response identification, (ii) validation timing detection, and (iii) validating response generation.
The emotional validation in dialogue systems can be decomposed into (i) validating response identification, (ii) validation timing detection, and (iii) validating response generation.
To support research on all three subtasks, we release \textbf{M-EDESConv}, a 120k English–Japanese multilingual corpus created through hybrid manual–automatic annotation, and \textbf{M-TESC}, a multilingual spoken-dialogue test set.
For timing detection, we propose \textbf{MEGUMI}, a Multilingual Emotion-aware Gated Unit for Mutual Integration, that fuses frozen XLM-RoBERTa semantics with language-specific emotion encoders via cross-modal attention and gated fusion.
% MEGUMI raises validation-precision to 51.1\% and macro F1 to 63.7\% on M-EDESConv, and remaining strongest on spontaneous speech, M-TESC with 41.4\% target precision.
MEGUMI shows superior performance on both the M-EDESConv and M-TESC datasets, both objectively and subjectively.
Finally, our \textbf{EmoValidBench} benchmarks of GPT-4.1 Nano and Llama-3.1 8B indicate that current LLMs generate contextually similar, diverse validating responses, but emotional understanding remains a major area for improvement. Project page: \url{https://github.com/zihaurpang/Multilingual-Emotional-Validation}
%\footnote{All code, data, and models will be released upon acceptance.}
\end{abstract}

\section{Introduction}\label{sec:1}

Empathy improves human--computer communication by fostering trust, rapport, and sustained engagement in human-robot interaction and conversational agents. Recent studies show that systems that modulate empathy in real time are more trusted and perceived as more helpful, underscoring the practical value of artificial empathy \cite{leite2013influence, morris2018towards, casas2021enhancing}. Accordingly, empathetic-dialogue research has enriched response generation with socio-cognitive signals. Representative directions include commonsense reasoning \cite{sabour2022cem, fu2023reasoning}, emotion-cause extraction \cite{gao2021improving}, and personality tailoring \cite{zhong2020towards, cai2024pecer, fu2024styemp}.

Yet generic ``I'm sorry to hear that'' replies often fall short for people who suppress emotions or face chronic stressors. Psychotherapy highlights \emph{\textbf{emotional validation}} - recognizing, understanding, and acknowledging others' emotional states, thoughts, and actions, as a deeper intervention that de-escalates negative affect and strengthens therapeutic alliance \cite{linehan1997validation}. Validating statements (e.g., ``It makes sense that you feel frustrated'') lower pain intensity in chronic-pain patients \cite{edlund2015see}, and foster adherence in youth mental-health care \cite{wasson2022youths} among other benefits \cite{carson2018effect, lambie2016role, daniel2023exploring}.

\begin{figure*}[t]
  \centering
  \includegraphics[width=\linewidth]{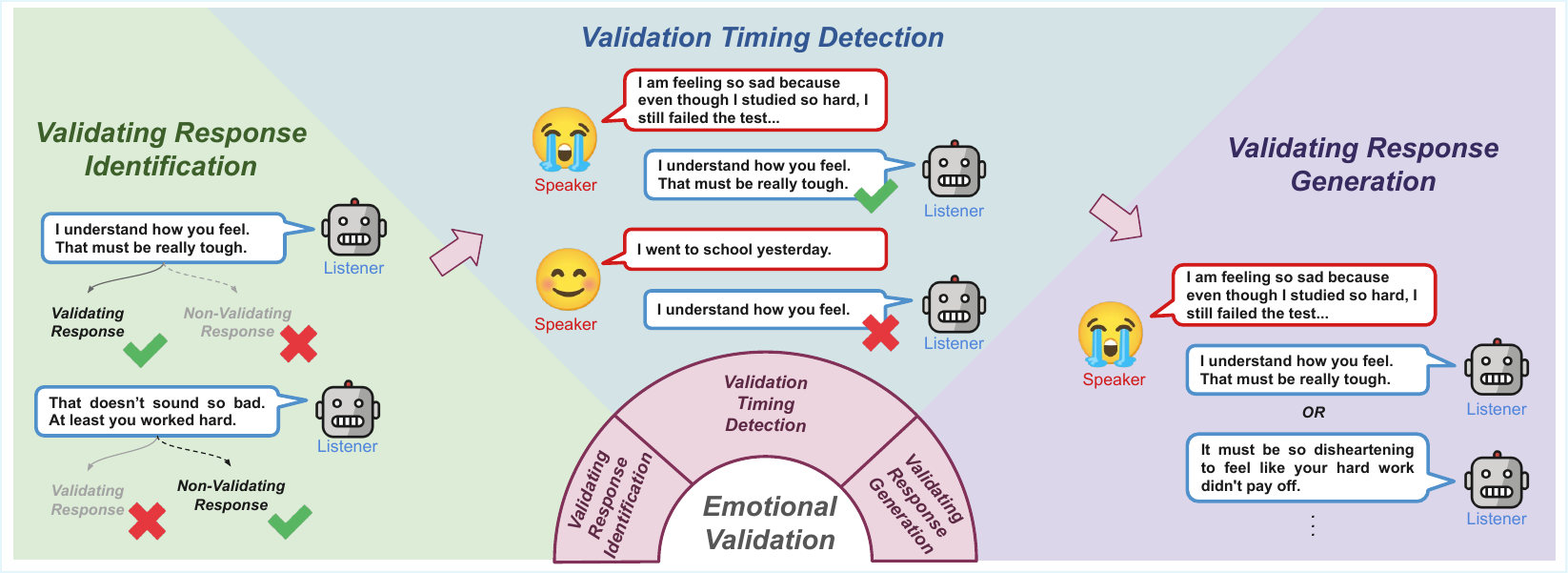}
  \caption{Emotional validation can be decomposed into three subtasks: (1) Validating Response Identification, which identifies \textit{what} constitutes a validating response; (2) Validation Timing Detection, which determines \textit{when} the system should validate the user; and (3) Validating Response Generation, which determines \textit{how} to generate an appropriate validating response.}
  \label{fig:framework}
\end{figure*}

Despite its importance in supportive interaction, computational work on emotional validation remains nascent. This gap is especially important in the era of Large Language Models (LLMs), as aligned LLMs have been reported to exhibit socially desirable and overly agreeable behavior, as well as sycophantic tendencies that favor agreement with the user even when such agreement is not well grounded \cite{sharma2023towards, wei2023simple, salecha2024large, bodrovza2024personality, hong2025measuring}. In emotional-support settings, this tendency can manifest as \emph{over-validation}: models may validate too early, too often, or without sufficient evidence that validation is warranted. Meanwhile, existing studies have largely relied on hand-crafted phrase lists and have been evaluated mainly in Japanese \cite{pang2023prediction, pang2024acknowledgment, pang2026paralinguistic}, leaving open how emotional validation should be modeled across languages whose emotion semantics, social-support norms, and listener feedback behavior may differ~\cite{jackson2019emotion, kim2008culture, clancy1996conversational}. 

To address these gaps, we formalize emotional validation in a multilingual dialogue framework, illustrated in Figure~\ref{fig:framework} and detailed in Appendix~\ref{sec:task_description}. Our main contributions are:

\begin{itemize}
    % \item To the best of our knowledge, we propose the first three-stage framework for emotional validation in dialogue system, that decomposes emotional validation into validating response identification, validation timing detection, and validating response generation, providing clear sub-tasks for future benchmarks. 
    \item We release \textbf{M-EDESConv}, and \textbf{M-TESC}, the first open-source, semi-automatic verified multilingual dialogue corpus annotated for validation phenomena in text and speech scenario, enabling cross-lingual evaluation beyond prior Japanese-only efforts. 
    % \item We introduce \textbf{MEGUMI}, which fuses monolingual emotional cues with multilingual semantic representations to detect validation timing more accurately. 
    \item We introduce \textbf{MEGUMI}, which fuses monolingual emotional cues with multilingual semantic representations to detect validation timing more accurately, helping reduce the over-validation tendency observed in LLMs.
    % \item \textbf{CoEV}: We present \emph{Chain-of-Emotional-Validation (CoEV)}, the first generation model that explicitly reasons over validation theory to decide what, why and how to acknowledge the user's emotional state.  
    % \item We present \textbf{EmoValidBench}, the first benchmark for validating response generation, evaluating the emotional validation of LLM baselines from semantic, diversity, empathy and LLM-as-judge metrics, to enable standardized comparisons across future models.
    \item We present \textbf{EmoValidBench}, the first benchmark for validating response generation, evaluating the emotional validation of LLM baselines from semantic, diversity, empathy and LLM-as-judge metrics, to enable standardized comparison not only of response quality but also of whether models validate appropriately rather than indiscriminately.
\end{itemize}

% % (see Subsection~\ref{eval_metrics}).
% % (i) semantic consistency (BLEU, BERTScore), (ii) lexical diversity (Distinct-1/2), and (iii) empathy-signal coverage (ER, IP, EX classifiers). 
% We report strong baselines using instruction-tuned large language models (GPT-4.1 nano, Llama-3.1 8B) under zero-shot, few-shot, and chain-of-thought prompting, which establishes a common test bed for future modeling efforts.

\section{Dataset Construction}

% We begin with two public English datasets emphasizing affective support. EmpatheticDialogues (ED) contains 24.8k two-speaker conversations collected from crowdworkers imagining specific emotional situations \cite{rashkin2018towards}. ESConv complements ED with 1,053 counselor-style dialogues in which trained volunteers comfort users facing real-life stressors \cite{liu2021towards}. For cross-lingual evaluation, we add Japanese ED \cite{sugiyama2021empirical} and create a Japanese ESConv via a GPT-4.1-mini\footnote{\href{https://openai.com/index/gpt-4-1/}{https://openai.com/index/gpt-4-1/}}
%  translation workflow. Combining English and Japanese versions yields our Multilingual-Empathetic Dialogue Emotional Support Conversation (M-EDESConv) dataset.

% To further evaluate under the spoken dialogue scenario, we include the TUT Emotional Storytelling Corpus (TESC), a Japanese two-party, multi-turn corpus (247 sessions; ≈9.2 h) where close friends recount experiences under eight Plutchik emotion prompts \cite{oishi2021design, plutchik2001nature}. We translate TESC to everyday English with the same workflow, yielding Multilingual-TUT Emotional Storytelling Corpus (M-TESC). Dataset summaries and prompts appear in Appendices \ref{sec:appendix}, \ref{sec:en-ja}, and \ref{sec:ja-en}. Translation quality, assessed with COMETKiwi \cite{rei2022cometkiwi}, are 0.8470/0.8479 for utterances/responses (0 = poor, 1 = perfect), indicating high overall quality for dataset construction, as detailed in Appendix \ref{sec:quality}.

We begin with two public English datasets emphasizing affective support, EmpatheticDialogues~\cite{rashkin2018towards}, and ESConv~\cite{liu2021towards}. To evaluate under the spoken dialogue scenario, we also include the TUT Emotional Storytelling Corpus (TESC)~\cite{oishi2021design}. Details of these dataset can be found in Appendix~\ref{sec:appendix}.

For cross-lingual evaluation, on top of adding a Japanese ED \cite{sugiyama2021empirical}, motivated by benchmarks like XNLI~\cite{conneau2018xnli} and PAWS-X~\cite{yang2019paws} that similarly adopted translation-based settings to obtain broad multilingual coverage, we create a Japanese ESConv via a GPT-4.1-mini\footnote{\href{https://openai.com/index/gpt-4-1/}{https://openai.com/index/gpt-4-1/}} translation workflow. Combining English and Japanese versions yields our Multilingual-Empathetic Dialogue Emotional Support Conversation (M-EDESConv) dataset. We also translate TESC to everyday English with the same workflow, yielding Multilingual-TUT Emotional Storytelling Corpus (M-TESC). Dataset summaries and  translation prompts appear in Appendices \ref{sec:appendix}, \ref{sec:en-ja}, and \ref{sec:ja-en}, respectively.

% Translation quality was assessed using both automatic and human evaluation. With COMETKiwi \cite{rei2022cometkiwi}, the translated utterances and responses obtained scores of 0.8470 and 0.8479, respectively, indicating high overall quality. We further conducted a human evaluation with three bilingual Japanese--English annotators, who rated the translations on accuracy, contextual consistency, emotional consistency, fluency/readability, and cultural appropriateness. The translations received an average score of 5.63/7 overall, with average fair level inter-annotator agreement (IAA) of 0.344, supporting the suitability of the translated data in this study. Full details are provided in Appendix \ref{sec:quality}.

Translation quality was assessed using both automatic and human evaluation. Using the COMETKiwi \cite{rei2022cometkiwi}, a reference-free quality estimation metric for machine translation, the translated utterances and responses obtained scores of 0.8470 and 0.8479, respectively, indicating high overall quality. Three bilingual Japanese--English annotators further evaluated the translations on accuracy, contextual consistency, emotional consistency, fluency/readability, and cultural appropriateness. The translations received an average score of 5.63/7 overall, with average fair inter-annotator agreement (IAA) of 0.344. Full details are provided in Appendix \ref{sec:quality}.

\begin{table}[htbp!]

\centering
\small
\begin{tabular}{lcc}
\hline
Dataset & \#Validation & \#Non-Validation \\
\hline
M-EDESConv & 46002 & 80551 \\
\textit{-train} & 36714 & 64540 \\
\textit{-val} & 4652 & 7867 \\
\textit{-test} & 4636 & 8144 \\
M-TESC & 1052 & 2028\\
\hline
\end{tabular}
\caption{Distribution of dataset} %in validation and non-validation
\label{tab:data_composition}
\end{table}

\subsection{Annotation of Emotional Validation}
\label{sec:3}

Given the impracticality of manually annotating all 120k utterances in our multilingual corpus, we adopted a two‐stage, hybrid annotation strategy inspired by large‐scale emotion datasets \citep{yang2012hybrid}. In the first stage, we manually labeled roughly 3000 utterances per source ($\approx$ 2\% of EmpatheticDialogue and 3\% of ESConv), classifying each response as either validating or non‐validating. This yielded 1204 validating and 1663 non‐validating responses in EmpatheticDialogue, and 680 validating versus 2258 non‐validating responses in ESConv. In the second stage, we trained a classifier to automatically label the remaining data. We frame this as a binary classification task using the response as input and the validation label as output, which is also known as \textbf{validating response identification} in this study. Through fine-tuning the \textit{xlm-roberta-large} model \cite{Alexis2019}, with training hyperparameters detailed in Appendix~\ref{sec:identify_para}, the classifier achieves a macro-average F1 score of 85.28 and an F1 score of 86.67 for the validation class on the manually annotated test set. The details of comparison to several baselines are presented in Appendix~\ref{sec:identify_result}, with result shown in Table~\ref{tab:automatic_annotation} .

Using the trained classifier, we proceed to annotate the full dataset. As a result, the final dataset made up of 98.6k utterances, with 5.89\% (5,805) of manual annotations, and 94.11\% (92,795 responses) were annotated automatically, detailed in Appendix~\ref{sec:dataset_details}. To verify the automatic labels' reliability, we ran an additional human evaluation, detailed in Appendix~\ref{sec:human_eval_auto}. With randomly 100 automatically annotated samples for each language, our crowdsourcing judgement result showed that 85\% of human judgement, with a substantial agreement level of 0.752 for the IAA, aligned with the automatic labels in the annotation process, showing the effectiveness of our hybrid annotation procedure in the dataset construction in this study. To further assess our task in a spoken dialogue setting, we additionally conduct manual annotation on the TESC dataset. Implementing the train/valid/test split with 8:1:1 ratio, the final distribution of validating and non-validating responses across datasets is summarized in Table~\ref{tab:data_composition}.

% %
% \begin{table*}[h]
% \caption{Results of validation timing detection task in multilingual [\%]}
% \label{tab:model_performance}
% \centering
% \begin{tabular}{lccccccc}
% \hline
% \multicolumn{1}{c}{} & \multicolumn{3}{c}{Macro Average} & & \multicolumn{3}{c}{Target Class} \\
% \cline{2-4}
% \cline{6-8}
% \multicolumn{1}{c}{} & Precision & Recall & F1-Score & & Precision & Recall & F1-Score \\ \hline
% Random Baseline & 50.20 & 50.21 & 49.23 & & 36.45 & 50.35 & 42.30 \\
% mBERT & 62.07 & 62.63 & 59.10 & & 45.16 & 74.15 & 56.19\\
% XLM-R & 62.78 & 63.13 & 59.03 & & 45.19 & 76.96 & 57.01\\
% Llama 3.1 8b &  &  &  & &  &  &  \\
% \textit{- Zero-shot} & 57.33 & 52.81 & 37.18 & & 37.72 & 93.75 & 53.79 \\
% \textit{- 3-shot} & 57.36 & 52.40 & 35.69 & & 37.49 & 94.84 & 53.73 \\
% GPT 4.1 Nano &  &  &  & &  &  &  \\
% \textit{- Zero-shot} & 58.42 & 57.74 & 51.68 & & 41.43 & 79.25 & 54.41 \\
% \textit{- 3-shot} & 58.87 & 56.39 & 46.75 & & 40.02 & 87.04 & 54.83 \\
% Proposed (Ours) &  &  &  & &  &  & \\

% \hline
% \end{tabular}
% \end{table*}
% %

% \subsection{Dataset Analysis}

\section{Validation Timing Detection}

\begin{figure*}[t]
  \centering
  \includegraphics[width=0.95\linewidth]{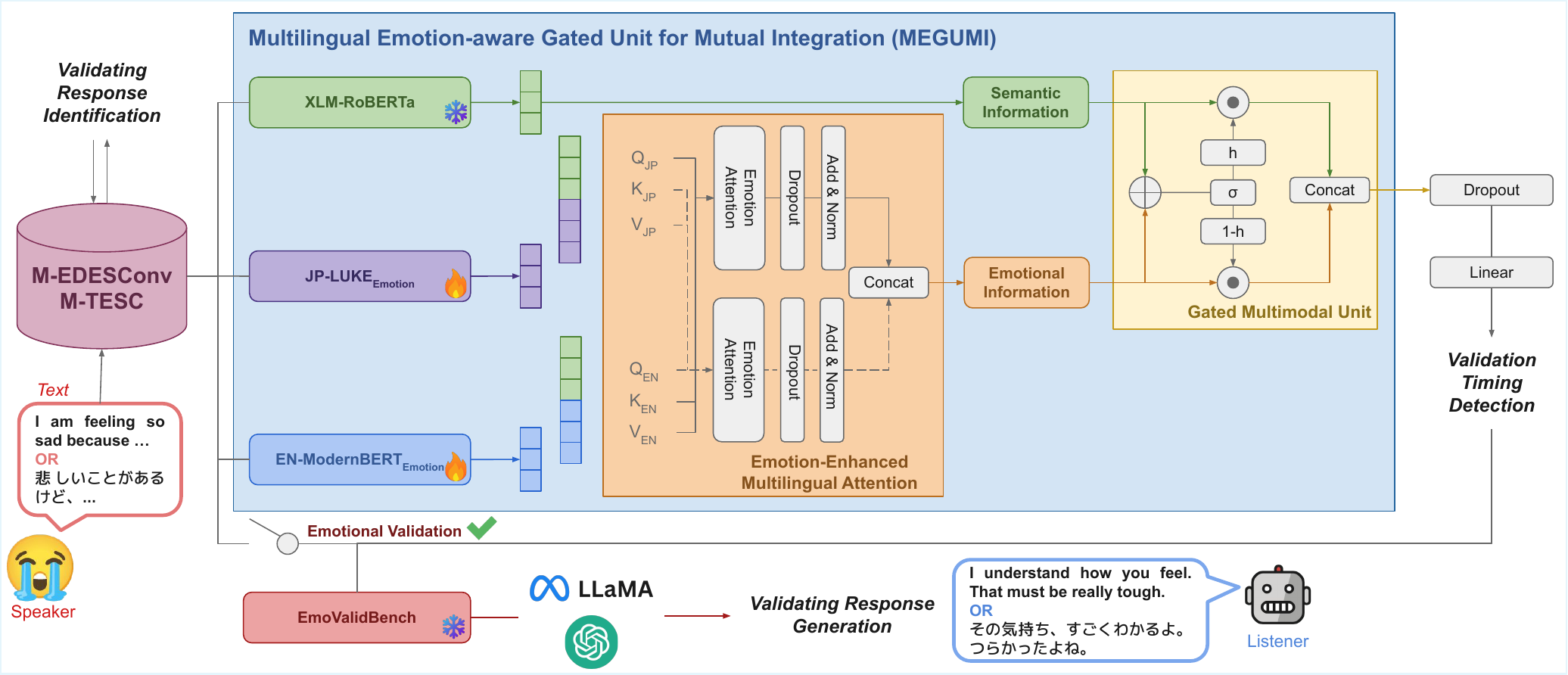} %\linewidth
  \caption{Overall proposed architecture in this study. We proposed a Multilingual Emotion-aware Gated Unit for Mutual Integration (MEGUMI) for the Validation Timing Detection Task. }
  \label{fig:architecture}
\end{figure*}

We cast validation timing detection as a binary classification problem: given the dialogue context up to the current user turn, decide whether the next system response should generate a validating response. Accurate timing requires both semantic content (what the user is saying) and affective cues (how they are feeling). The proposed Multilingual Emotion-aware Gated Unit for Mutual Integration (MEGUMI) architecture, shown in Figure~\ref{fig:architecture}, fuses language-agnostic semantics with language-specific emotion representations through a gated, cross-modal pipeline that can be trained end-to-end from text utterance alone.

\subsection{Validation Timing Detection Modal}

\paragraph{Semantic backbone}The core encoder of MEGUMI is XLM-RoBERTa-large\footnote{\href{https://huggingface.co/FacebookAI/xlm-roberta-large}{https://huggingface.co/FacebookAI/xlm-roberta-large}} , chosen for its strong zero-shot transfer across 100+ languages. We freeze its parameters to preserve multilingual lexical knowledge and to curb computational cost.

\paragraph{Language-specific emotion channels}Research shows that emotion taxonomies and lexical cues vary by language; a single encoder therefore risks conflating culture-specific signals \cite{takenaka2025performance}. For English utterances, we leverage ModernBERT-large\footnote{\href{https://huggingface.co/cirimus/modernbert-large-go-emotions}{https://huggingface.co/cirimus/modernbert-large-go-emotions}} fine-tuned on GoEmotions - a 58 k-instance Reddit corpus with 27 fine-grained labels \cite{demszky-etal-2020-goemotions}. Japanese turns are processed by LUKE-Japanese-large adapted to the WRIME writer-emotion dataset\footnote{\href{https://huggingface.co/Mizuiro-sakura/luke-japanese-large-sentiment-analysis-wrime}{https://huggingface.co/Mizuiro-sakura/luke-japanese-large-sentiment-analysis-wrime}}. 
% The emotion [CLS] vector from the relevant channel is concatenated with the frozen semantic [CLS]. \textbf{FORMULA}

% In our model, z denotes the intermediate embedding vector output for each modality. Specifically:

% $z_{semantic}$ represents the output from the semantic encoder (XLM-RoBERTa).

% $z_{emotion}$ represents the output from the emotion encoders ($LUKE_{Emotion}$ for Japanese and $ModernBERT_{Emotion}$ for English).

% These embeddings are then fused through our Gated Multimodal Unit (GMU), which learns an optimal weighting between semantic and emotional information for the validation timing detection task.

% yielding a 2 048-d representation per language.

\paragraph{Emotion-Enhanced Multilingual Attention (EEMA)} In each training batch, we ensure that both English and Japanese emotional cues are present and processed jointly. To capture their interactions, we introduce an emotion-enhanced multilingual attention block inspired by the Multimodal Transformer~\cite{tsai2019multimodal}. Specifically, the module projects the concatenated representation from one language as the query and the representation from the other language as the key and value, then applies scaled dot-product attention and returns two residual-normalized streams. With both English and Japanese emotion channels available during training, EEMA allows MEGUMI to learn latent cross-lingual alignments between affective cues  expressed differently across languages (e.g., fear in English vs. anger in Japanese).

% which uses directional pairwise cross-modal attention to model interactions across streams 
% \paragraph{Emotion-Enhanced Multilingual Attention (EEMA)} In each training batch, we ensure both English and Japanese emotional cues are present to be processed simultaneously. we introduce an emotion-enhanced multilingual attention block inspired by the Multimodal Transformer \cite{tsai2019multimodal}. The module projects one lingual's concatenated vector as query and the other as key/value, computes scaled dot-product attention, and returns two residual-normed streams. With the existence of both Japanese and English emotion channels, it allows MEGUMI to learn latent alignments for affective cues (e.g., fear in English vs. anger in Japanese) across the languages simultaneously. 
% In the monolingual setting, EEMA becomes the single-stream special case of the same module, which is equivalent to self-attention over the available emotional representation.

% semantic patterns (e.g., ``I lost my job'') and 

\paragraph{Gated Multimodal Unit}Simply concatenating streams can swamp minority cues; we therefore integrate them through a Gated Multimodal Unit, which has proven effective in image–text genre classification \cite{arevalo2017gated}. A sigmoid gate 
$h$ decides, per sample, how much of the emotion-projected vector, $z_{emotion}$
versus the semantic-projected vector, $z_{semantic}$ to pass forward:
% , producing a 768-d fused representation: 
% \[
$z=\mathbf{h}z_{semantic}+\mathbf{(1-h)}z_{emotion}$.
% \]
The fused vector is fed to a dropout–linear softmax head for the binary labels validate or non-validate. To counter class imbalance we weight the cross-entropy loss.
% by inverse-frequency factors computed from the training split. 
% Reproducibility is ensured via deterministic seeds, and all components except the text encoder are fine-tuned.

\subsection{Experimental Settings}
We fine-tuned all models on the M-EDESConv corpus, with hyperparameters detailed in Appendix \ref{sec:setting_vtd}. We benchmarked against (i) a random classifier, (ii) two fine-tuned multilingual language models, mBERT and XLM-RoBERTa \cite{Alexis2019}, and (iii) instruction-tuned large language models: Llama-3.1.1 8B-Instruct and GPT-4.1 nano, each in zero-shot, 3-shot prompt-engineering, CoT and LoRA regimes, refer to Appendix~\ref{sec:vtd}. 
% Larger 70B Llama variants were excluded owing to real-time memory constraints. 

\subsection{Ablation Studies}
To disentangle architectural choices we compared: 

\paragraph{XLM-RoBERTa} uses only the frozen semantic encoder, followed by dropout and a linear layer.

\paragraph{+Mono-EN} uses only the English Emotion Embedding (EE) Encoder, fused with XLM-RoBERTa output, followed by dropout and a linear layer.

\paragraph{+Mono-JP} uses the Japanese Emotion Embedding Encoder in the same fashion.

\paragraph{+Multi-Concat} combines both English and Japanese EE Encoders via direct concatenation, together with the XLM-RoBERTa output.

\paragraph{+Multi-EEMA} feeds the outputs of the English and Japanese EE Encoders into the EEMA module to compute a single, fused emotional representation, which is then concatenated with the semantic encoder's output before classification.

\begin{table*}[h]
\centering
\scriptsize
\begin{tabular}{lcccc c cccc c cccc}
\toprule
                & \multicolumn{4}{c}{M-EDESConv} & & \multicolumn{4}{c}{M-TESC} & & \multicolumn{4}{c}{Human-validated subset} \\ \cline{2-5} \cline{7-10} \cline{12-15}
                & \multicolumn{1}{c}{Overall} & \multicolumn{3}{c}{Target Class} & & \multicolumn{1}{c}{Overall} & \multicolumn{3}{c}{Target Class} & & \multicolumn{1}{c}{Overall} & \multicolumn{3}{c}{Target Class} \\ \cline{2-5} \cline{7-10} \cline{12-15}
                & M-F1 & Pre. & Rec. & F1 & & M-F1 & Pre. & Rec. & F1 & & M-F1 & Pre. & Rec. & F1 \\ 
\midrule
\multicolumn{14}{c}{\textit{Multilingual}}
\\
\midrule
Random Baseline      & 49.23 & 36.45 & 50.35 & 42.30 & & 49.03 & 34.41 & 50.02 & 40.77 & & 48.68      &  48.71     &  47.60     &  48.13     \\
mBERT                & \underline{59.10} & 45.16 & 74.15 & 56.14 & & 53.29 & 38.29 & 41.35 & 39.76 & & 58.80 & 60.47 & 52.00 & 55.91 \\
XLM-RoBERTa          & 59.03 & \underline{45.19} & 76.96 & \underline{56.94} & & \underline{55.10} & \underline{40.24} & 50.00 & 44.60 & & 58.85 & 60.23 & 53.00 & 56.38 \\

Llama 3.1 8b         &       &       &       &       & &       &       &       &       & &       &       &       &       \\
\textit{- Zero-shot} & 37.18 & 37.72 & \underline{93.75} & 53.79 & & 40.76 & 36.31 & \underline{89.58} & \underline{51.68} & &  36.54     & 50.16      &  96.80     &  66.07     \\
\textit{- 3-shot}    & 35.69 & 37.49 & \textbf{94.84} & 53.73 & & 31.53 & 34.81 & \textbf{96.29} & 51.13 & & 39.12 & 51.04 & 98.00 & 67.12 \\
\textit{- LoRA}      & 34.42 & 36.40 & 90.81 & 51.97 & & 47.99 & 33.83 & 58.44 & 42.86 & &  42.02     &  51.85     &  98.00     &  67.82     \\

GPT 4.1 Nano         &       &       &       &       & &       &       &       &       & &       &       &       &       \\
\textit{- Zero-shot} & 51.68 & 41.43 & 79.25 & 54.41 & & 54.04 & 39.95 & 66.46 & 49.90 & & 51.76      & 52.14      &  46.80     & 49.31 \\
\textit{- 3-shot}    & 46.75 & 40.02 & 87.04 & 54.99 & & 50.14 & 39.01 & 80.93 & \textbf{52.65} & & 49.50 & 53.01 & 88.00 & 66.17       \\
\textit{- CoT}    &  42.19 & 39.16 & 92.88 & 55.09 && 44.16 & 36.19 & 86.88 & 51.09 && 46.54 & 52.89 & 95.40 & 68.05     \\
MEGUMI (Ours)       & \textbf{63.71}$^{\dagger}$  & \textbf{51.07} & 66.11 & \textbf{57.62} & & \textbf{56.89}$^{\dagger}$  & \textbf{41.44} & 48.70 & 44.78 & & 61.15 & 64.20 & 52.00 & 57.46 \\
Human                & \multicolumn{4}{c}{No Applicable}  & & \multicolumn{4}{c}{No Applicable} & & 70.82 & 66.20 & 88.67 & 75.78 \\ 
\midrule
\multicolumn{14}{c}{\textit{Monolingual}}
\\
\midrule
\rowcolor[gray]{0.95}\textbf{English}    &  &  &  &  &  &  &  &  &  &  &  & & & \\
Random Baseline & 49.40 & 35.83 & 50.59 & 41.95 & & 48.84 & 34.29 & 50.27 & 40.76 && 47.00 & 47.17 & 50.00 & 48.54\\
BERT            & 59.74 & \underline{45.16} & 74.15 & 56.19 & & 53.29 & 38.29 & 41.35 & 39.76 && 63.47 & 61.29 & 76.00 & 67.86\\
ModernBERT      & 59.10 & 44.94 & 75.02 & \underline{56.27} & & 52.80 & 37.67 & 46.77 & 41.73 && 69.81 & 67.24 & 78.00 & 72.22 \\
% mBERT & -- & -- & -- & -- & & -- & -- & -- & -- && 69.66 & 86.21 & 50.00 & 63.29 \\
XLM-RoBERTa     & \textbf{62.22} &  47.24 & 70.38 & \textbf{56.57} & & \underline{54.83} & \underline{39.94} & 52.09 & 45.21 && 69.00 & 73.17 & 60.00 & 65.93 \\
Llama 3.1 8b    &  &  &  &  &  &  &  &  &  &  &  \\
\textit{- Zero-shot} & 40.90 & 37.55 & 90.54 & 53.08 & & 49.07 & 38.55 & 81.64 & \textbf{52.37} && 37.57 & 49.68 & 95.20 & 65.28  \\
\textit{- 3-shot}    & 33.75 & 36.45 & \textbf{96.47} & 52.91 & & 34.15 & 35.41 & \textbf{95.93} & \underline{51.73} && 42.75 & 51.60 & 97.60 & 67.51 \\
\textit{- LoRA}    & 50.18 & 33.33 & 31.17 & 32.21 && 50.18 & 33.33 & 31.17 & 32.21 && 44.16 & 50.59 & 86.00 & 63.70\\
GPT 4.1 Nano         &  &  &  &  &  &  &  &  &  &  &  \\
\textit{- Zero-shot} & 55.36 & 42.53 & 74.43 & 54.13 & & 53.74 & 39.29 & 60.80 & 47.73 && 50.02 & 50.69 & 46.00 & 48.21 \\
\textit{- 3-shot}    & 56.69 & 43.19 & 68.49 & 52.98 & & 53.70 & 39.24 & 60.57 & 47.62 && 49.11 & 51.97 & 84.40 & 64.33 \\
\textit{- CoT}  & 43.84 & 38.88 & \underline{92.39} & 54.73 && 46.66 & 36.53 & \underline{82.08} & 50.56 && 47.31 & 53.13 & 95.20 & 68.20\\
MEGUMI (Ours)        & \underline{61.67} & \textbf{48.40} & 60.99 & 53.97 & & \textbf{58.41} & \textbf{45.07} & 41.56 & 43.24 && 64.58 & 76.67 & 46.00 & 57.50 \\
Human                & \multicolumn{4}{c}{No Applicable}  & & \multicolumn{4}{c}{No Applicable} & & 70.81 & 66.84 & 84.67 & 74.71 \\

\rowcolor[gray]{0.95}\textbf{Japanese}   &  &  &  &  &  &  &  &  &  &  & & & & \\
Random Baseline & 49.01 & 37.23 & 50.09 & 42.71 & & 49.22 & 34.54 & 49.77 & 40.77 && 47.00 & 47.17 & 50.00 & 48.54\\
BERT            & 57.35 & 44.83 & 76.57 & 56.61 & & 55.67 & \underline{40.90} & 58.56 & 48.16 && 53.70 & 53.45 &62.00 &57.41 \\
ModernBERT      & 51.15 & 42.02 & 86.48 & \underline{56.67} & & 51.51 & 38.18 & 66.92 & 48.62 && 61.80 & 59.46 & 88.00 & 70.97\\
% mBERT & -- & -- & -- & -- & & -- & -- & -- & -- && 53.70 & 53.45 & 62.00 & 57.41\\
XLM-RoBERTa     & \underline{60.76} & \underline{47.52} & 69.73 & 56.56 & & \underline{54.39} & 39.45 & 49.05 & 43.73 && 55.93 & 55.56 & 60.00 & 57.69 \\
Llama 3.1 8b    &  &  &  &  &  &  &  &  &  &  &  \\
\textit{- Zero-shot} & 30.98 & 37.79 & 97.82 & 54.51 & & 29.75 & 34.74 & \underline{98.21} & 51.33 && 35.36 & 50.62 & 98.40 & 66.85 \\
\textit{- 3-shot}    & 29.22 & 37.69 & \underline{99.44} & 54.67 & & 27.07 & 34.46 & \textbf{99.85} & 51.23 && 38.83 & 51.57 & 98.80 & 67.77 \\
\textit{- LoRA}    & 38.71 & 34.02 & 85.71 & 48.71 && 38.71 & 34.02 & 85.71 & 48.71 && 37.63 & 52.02 & 98.50 & 68.08 \\
GPT 4.1 Nano         &  &  &  &  &  &  &  &  &  &  &  \\
\textit{- Zero-shot} & 47.03 & 40.28 & 81.51 & 53.91 & & 52.48 & 39.63 & 74.68 & \underline{51.78} & & 53.45 & 53.60 & 47.60 & 50.40 \\
\textit{- 3-shot}    & 39.67 & 39.28 & \textbf{99.94} & 55.22 & & 43.38 & 37.53 & 91.41 & \textbf{53.21} && 44.93 & 52.29 & 92.00 & 66.67 \\
\textit{- CoT}    & 40.02 & 39.47 & 93.43 & 55.50 && 41.28 & 35.88 & 91.69 & 51.57 && 45.74 & 52.65 & 95.60 & 67.90\\
MEGUMI (Ours)        & \textbf{61.20} & \textbf{48.83} & 75.87 & \textbf{59.42} & & \textbf{57.09} & \textbf{41.35} & 55.84 & 47.51 && 57.00 & 56.86 & 58.00 & 57.43 \\
Human                & \multicolumn{4}{c}{No Applicable}  & & \multicolumn{4}{c}{No Applicable} & & 70.75 & 65.57 & 92.67 & 76.80  \\

\bottomrule
\end{tabular}
\caption{Results of validation timing detection task [\%]. Top two results are highlighted in \textbf{bold} and \underline{underline}, respectively. $^{\dagger}$ indicates a statistically significant difference for $p < 0.05$ between MEGUMI and the best baselines, XLM-RoBERTa, determined by t-test.}
\label{tab:timing_detection}
\end{table*}

\subsection{Evaluation Metrics}
% In the context of everyday conversation, when a system chooses to validate matters more than how often it does so. Consequently, we treat target-class precision - the proportion of predicted validate turns that truly warrant validation - as the principal metric. 
A model that indiscriminately labels many turns as validating (high recall) risks producing hollow or repetitive acknowledgments that undermine perceived empathy; hence a high F1 score alone can be misleading if it masks low precision. We therefore report (i) validation-precision as the primary indicator of conversational appropriateness, (ii) validation-F1 to capture the precision-recall trade-off, and (iii) macro-F1 (M-F1) across both classes to ensure that performance on the majority non-validate class is not neglected.

\subsection{Results and Discussion}

Table~\ref{tab:timing_detection} summarizes the results on validation-timing detection in multilingual and monolingual settings. In the multilingual setting, \textsc{MEGUMI} performs best overall: it achieves the highest macro-F1, target-class precision, and target-class F1 on M-EDESConv, and again the best macro-F1 and target-class precision on M-TESC, showing strong generalization to spoken dialogue. Across model types, LLM-based methods consistently trade precision for recall, often over-predicting validation timing, whereas fine-tuned classifiers are better balanced; \textsc{MEGUMI} improves this balance further. The monolingual results show the same overall trend with some dataset-specific differences. On M-EDESConv, fine-tuned PLMs remain strongest overall, with XLM-RoBERTa achieving the best macro-F1 in both languages, while \textsc{MEGUMI} yields the highest target-class precision in both languages. On M-TESC, however, \textsc{MEGUMI} achieves the best macro-F1 and target-class precision in both languages, indicating better calibration under noisier spoken conditions. Although some LLM baselines obtain the highest target-class F1 on M-TESC due to extremely high recall, their lower precision suggests more false positives.

Ablation study in Table~\ref{tab:ablation} shows that multilingual emotion information plays an important role. \textit{+Multi-Concat} outperforms both \textit{+Mono-EN} and \textit{+Mono-JP}, suggesting that English and Japanese emotion cues are complementary. Adding Emotion-Enhanced Multilingual Attention (\textit{+Multi-EEMA}) yields further gains, and the full model performs best overall. These results indicate that cross-lingual affective cues, beyond lexical semantics alone, help identify validation-worthy moments, especially in emotionally ambiguous contexts.

\begin{table}[htbp!]
\centering
\scriptsize
\resizebox{\linewidth}{!}{%
\begin{tabular}{lccccc}
\hline
\multicolumn{1}{c}{} & \multicolumn{1}{c}{Overall} & & \multicolumn{3}{c}{Target Class} \\
\cline{2-2}
\cline{4-6}
\multicolumn{1}{c}{} & M-F1 & & Pre. & Rec. & F1 \\ \hline
XLM-RoBERTa      & 47.29 & & 39.60 & 82.42 & 53.50 \\
+ Mono-EN        & 57.27 & & 45.10 & 48.23 & 46.61 \\
+ Mono-JP        & 56.86 & & 43.97 & 62.88 & 51.75 \\
+ Multi-Concat   & 59.80 & & 46.75 & 73.86 & 57.26 \\
+ Multi-EEMA     & 62.48 & & 49.70 & 65.31 & 56.45 \\
MEGUMI (Ours)    & 63.71 & & 51.07 & 66.11 & 57.62 \\
\hline
\end{tabular}
}
\caption{Ablation study result [\%].}
% , showing the impact of different model variants on macro-average F1-score and target-class precision, recall, and F1-score 
\label{tab:ablation}
\end{table}

To further examine task validity, we conducted an additional human evaluation on 200 utterances (100 English and 100 Japanese), balanced across validate and non-validate labels. Three native speakers judged whether a validating response should follow each utterance. The automatic labels achieved 72\% average agreement with human judgments, with per-annotator agreement of 69--73\% for English and 70--75\% for Japanese. The overall IAA was moderate ($\kappa=0.448$), suggesting that validation timing is more subjective than response-level validation labeling but still sufficiently consistent to support the task. On this human-validated subset, shown in the rightmost part of Table~\ref{tab:timing_detection}, \textsc{MEGUMI} again achieves the best macro-F1 among automatic systems and the highest target-class precision, outperforming mBERT, XLM-RoBERTa, and the LLM baselines.

\section{Validating Response Generation}

We position validating response generation as a stand-alone benchmark task, with the introduction of \textbf{EmoValidBench}, that tests whether a system can produce a concise, theory-consistent acknowledgment once the dialogue context has been flagged as requiring validation. 
% This section details the benchmark design, experimental protocol, automatic metrics, and baseline results.

\subsection{Benchmark construction}
From the M-EDESConv corpus we extract every user utterance whose gold timing label is validate. Each of these turns is paired with one or more human validating replies that serve as references. English inputs are pre-processed with the Moses tokenizer\footnote{\href{https://github.com/luismsgomes/mosestokenizer}{https://github.com/luismsgomes/mosestokenizer}}, while Japanese inputs are segmented by MeCab + UniDic\footnote{\href{https://taku910.github.io/mecab/}{https://taku910.github.io/mecab/}} to ensure comparability across BLEU and Distinct-n implementations.

We prompted Llama-3.1 8B and GPT-4.1 nano in Zero-shot (only the task definition), 3-shot (three labelled dialogue exemplars per language), Zero-shot CoT (``Let's think step by step'' preamble) \cite{kojima2022large}, and LoRA-based fine-tuning~\cite{hu2022lora}, see Appendix~\ref{sec:vrg}. 
\begin{table*}[h]
\centering
\scriptsize
\resizebox{\textwidth}{!}{%
\begin{tabular}{lcc c cc c ccc c c ccccccc}
\toprule
& \multicolumn{8}{c}{Traditional LM-based metrics} & & \multicolumn{6}{c}{LLM-as-Judge} \\
\cline{2-9} \cline{11-16}
\multicolumn{1}{c}{} & BERT & BLEU & D1 & D2 & ER & IP & EX & AVG & & AF & AR & RN & SW & SB & Ov. \\ 
\midrule
\multicolumn{16}{c}{\textit{Multilingual}}
\\
\midrule
Llama 3.1 8b &  &  &  &  &  &  &  &  & &  &  &  &  &  &  \\
\textit{- Zero-shot} & 88.67 & 15.46 & 4.88 & 23.96 & 52.90 & 70.94 & 67.14 & 46.28 & & 4.02 & 4.90 & 6.11 & 4.68 & 7.00 & 65.98 \\
\textit{- 3-shot}    & 89.03 & 15.94  & 5.52 & 25.97 & 54.66 & 71.54 & 69.59 & 47.61 & & 4.42 & 5.06 & 6.65 & 4.96 & 7.00 & 68.96 \\
\textit{- LoRA}      & 89.29 & 15.20  & 12.05 & 30.60 & 59.52 & 65.36 & 64.43 & 48.06 & & 4.45 & 5.21 & 6.69 & 5.30 & 7.00 & 69.55 \\

GPT 4.1 Nano &  &  &  &  &  &  &  &  & &  &  &  &  &  &  \\
\textit{- Zero-shot} & 88.87 & 13.47 & 4.80 & 22.37 & 51.95 & 73.17 & 69.97 & 46.37 & & 4.96 & 5.22 & 6.73 & 5.41 & 7.00 & 71.69 \\
\textit{- 3-shot}    & 89.26 & 16.24  & 4.58 & 21.42 & 52.32 & 72.29 & 68.76 & 46.41 & & 5.05 & 5.57 & 6.90 & 5.68 & 7.00 & 73.57 \\
\textit{- CoT}       & 88.13 & 17.71 & 4.83 & 24.53 & 50.72 & 73.30 & 71.21 & 47.21 & & 5.12 & 5.68 & 6.92 & 5.86 & 7.00 & 75.10 \\

\midrule
\multicolumn{16}{c}{\textit{Monolingual}} \\
\midrule
\rowcolor[gray]{0.95}\textbf{English} & & & & & & & & & & & & & & & \\
Llama 3.1 8b &  &  &  &  &  &  &  &  & &  &  &  &  &  &  \\
\textit{- Zero-shot} & 88.36 & 13.20 & 4.35 & 23.94 & 62.37 & 76.11 & 68.73 & 48.15 & & 5.34 & 4.75 & 6.86 & 5.39 & 6.98 & 61.33 \\
\textit{- 3-shot}    & 88.45 & 13.17 & 4.51 & 24.93 & 62.01 & 76.73 & 67.59 & 48.20 & & 6.05 & 5.52 & 6.97 & 5.99 & 6.99 & 67.12  \\
\textit{- LoRA}    & 89.30 & 12.01 & 7.53 & 30.60 & 65.13 & 74.73 & 71.74 & 50.06 &  & 3.66 & 3.18 & 6.31 & 4.32 & 6.96 & 63.25 \\

GPT 4.1 Nano &  &  &  &  &  &  &  &  & &  &  &  &  &  &  \\
\textit{- Zero-shot} & 88.33 & 12.86 & 5.02 & 25.60 & 62.73 & 75.65 & 69.10 & 48.47 & & 5.95 & 5.33 & 6.94 & 5.80 & 6.99 & 74.24 \\
\textit{- 3-shot}    & 88.53 & 13.32 & 4.59 & 22.88 & 60.71 & 75.52 & 67.75 & 47.61 & & 5.68 & 5.06 & 6.95 & 5.77 & 6.99 & 74.62 \\
\textit{- CoT}       & 88.13 & 12.78 & 4.77 & 27.01 & 62.37 & 75.79 & 67.86 & 48.39 & & 5.97 & 5.41 & 6.94 & 5.90 & 6.99 & 74.69 \\

\rowcolor[gray]{0.95}\textbf{Japanese} & & & & & & & & & & & & & & & \\
Llama 3.1 8b &  &  &  &  &  &  &  &  & &  &  &  &  &  &  \\
\textit{- Zero-shot} & 89.15 & 18.24 & 5.23 & 23.67 & 54.73 & 72.07 & 61.38 & 46.35 & & 3.93 & 3.35 & 6.38 & 4.04 & 7.00 & 59.79 \\
\textit{- 3-shot}    & 89.55 & 19.92 & 5.79 & 24.97 & 54.67 & 74.27 & 61.10 & 47.18 & & 4.37 & 3.77 & 6.33 & 4.42 & 6.99 & 64.63 \\
\textit{- LoRA}    & 89.34 & 16.56 & 3.39 & 15.02 & 56.92 & 75.38 & 58.84  & 45.06 && 3.03 & 2.70 & 5.66 & 3.29 & 6.95 & 56.32\\

GPT 4.1 Nano &  &  &  &  &  &  &  &  & &  &  &  &  &  &  \\
\textit{- Zero-shot} & 89.54 & 19.96 & 4.56 & 18.40 & 53.34 & 76.77 & 61.62 & 46.31 & & 5.49 & 4.90 & 6.94 & 5.58 & 7.00 & 73.42 \\
\textit{- 3-shot}    & 90.16 & 23.16 & 5.00 & 20.33 & 57.70 & 77.60 & 60.83 & 47.83 & & 5.61 & 4.99 & 6.95 & 5.63 & 7.00 & 74.24 \\
\textit{- CoT}       & 88.14 & 14.38 & 7.00 & 27.74 & 49.90 & 78.09 & 61.35 & 46.66 & & 5.72 & 5.16 & 6.93 & 5.73 & 7.00 & 74.30 \\
\bottomrule
\end{tabular}%
}
\caption{Validating response generation results [\%]. The left block reports traditional language-model-based metrics. The right block reports multilingual LLM-as-Judge scores. AF = Acknowledges Feelings, AR = Accurate Reflection, RN = Respect / Nonjudgment, SW = Support / Warmth, SB = Safety / Boundaries, and Ov. = Overall.}
\label{tab:response_generation}
\end{table*}

\subsection{Evaluation metrics}
\label{eval_metrics}
To comprehensively assess validating response generation, we employ traditional-language-model-based metrics that capture semantic consistency, lexical diversity, and empathetic signal presence, as well as more-modern LLM-as-judge-based metrics~\cite{zheng2023judging}, that captures several clinical-grounded metrics. We also employ a small-scale of human evaluation on top of objective evaluation metrics.

\subsubsection{Traditional-Language-Model Evaluation}

\paragraph{Semantic Consistency.} We utilize BERTScore \cite{zhang2019bertscore} and BLEU~\cite{papineni2002bleu} to evaluate the semantic alignment between generated responses and reference texts. 
% BERTScore leverages contextual embeddings to assess deeper semantic correspondence, reporting precision, recall, and F1 scores.
\paragraph{Lexical Diversity.} To quantify the diversity of generated language, we calculate Distinct-1 (D1) and Distinct-2 (D2)~\cite{li2015diversity}, which represent the ratios of unique unigrams and bigrams to the total number of tokens. 
% Higher values indicate a broader range of lexical choices, reflecting more varied and potentially more engaging responses. 
\paragraph{Empathetic Signal Coverage.} Inspired by prior work on empathetic communication~\cite{sharma2020computational, lee2022does, fu2024styemp}, we introduce three categories of empathetic signals: IP (interpretations), EX (explorations), and ER (emotional reactions). Specifically, IP denotes acknowledgments and understanding of, the interlocutor's emotion or situation; EX denotes expressions of active interest in the interlocutor's situation; and ER denotes explicit emotional reactions. The training details are shown in Appendix~\ref{sec:empathytrain}, while the evaluation results of the empathetic-signal predictors are summarized in Table~\ref{tab:empathyaccu}.

\subsubsection{LLM-as-judge-based Evaluation}

We further complement the objective evaluation with a structured LLM-as-judge rubric grounded in clinical communication frameworks, including NURSE~\cite{childers2023beyond} and OARS~\cite{rollnick1995motivational}. Our rubric operationalizes whether the response (i) acknowledges feelings, (ii) reflects the user's experience accurately, (iii) maintains respect and a non-judgmental tone, (iv) conveys warmth and support, and (v) preserves safety and appropriate conversational boundaries. The complete prompt shown in the Appendix~\ref{sec:llmjudgeprompt}.

\subsubsection{Subjective Human Evaluation}

To directly assess therapeutic appropriateness, we additionally conduct a blinded human evaluation on 100 utterance--response pairs (50 English, 50 Japanese). Each item is rated by three native speakers recruited through crowdsourcing. We evaluate six dimensions: \textit{Naturalness}, \textit{Contextual Understanding}, \textit{Emotional Understanding}, \textit{Non-judgmental Tone}, \textit{Mindful Presence}, and \textit{Perceived Validation}. These dimensions are designed to capture whether a response feels emotionally appropriate and validating beyond surface semantic overlap.

\begin{table}[t]
\centering
\small
\resizebox{\linewidth}{!}{%
\begin{tabular}{lccc}
\toprule
Criterion & Ground Truth & Llama 3.1 8B & GPT-4.1 Nano \\
\midrule
Naturalness & 5.48 & 5.27 & 5.42 \\
Contextual Understanding & 5.76 & 5.67 & 5.67 \\
Emotional Understanding & 5.92 & 5.35 & 5.59 \\
Non-judgmental Tone & 6.08 & 5.98 & 5.85 \\
Mindful Presence & 5.66 & 5.53 & 5.43 \\
Perceived Validation & 5.71 & 5.33 & 5.47 \\
% \textit{Average} & 5.77 & 5.52 & 5.57 \\
\midrule
IAA ($\alpha$) & 0.406 & 0.453 & 0.322 \\
\bottomrule
\end{tabular}
}
\caption{Human evaluation of validating-response quality for ground truth and 3-shot prompted model outputs. Three annotators rated each response on six dimensions; IAA denotes Krippendorff's $\alpha$.}
\label{tab:human_eval_generation}
\end{table}

\subsection{Results and Discussion}
\label{sec:response_result}

Table~\ref{tab:response_generation} reports five-seed averages for validating-response generation in multilingual and monolingual settings. Overall, compact LLMs produce contextually appropriate responses that are semantically close to the references, but remain weaker in affective quality. This trend is consistent across both traditional metrics and LLM-as-Judge scores: models score relatively high on \textit{Respect/Nonjudgment} and \textit{Safety/Boundaries}, but lower on \textit{Acknowledges Feelings}, \textit{Accurate Reflection}, and \textit{Support/Warmth}. The same pattern appears in the automatic empathy metrics, where models are stronger at showing interest in the speaker's situation (IP) than at recognizing emotion (ER) or expressing explicit emotional reaction (EX). Overall, current LLMs appear generally polite and safe, but less reliable at explicit emotional validation.

Across settings, GPT~4.1 Nano consistently outperforms Llama~3.1~8b in judged overall quality. In the multilingual setup, GPT with CoT achieves the best overall score (75.10), and the same tendency holds in both monolingual languages. Although Japanese often matches or exceeds English on surface-oriented metrics such as BERTScore and BLEU, this does not yield better judged validation quality. More broadly, English performs better than Japanese on emotion-sensitive criteria, especially \textit{Acknowledges Feelings}, \textit{Accurate Reflection}, and \textit{Support/Warmth}. We attribute this gap to English-dominant training and evaluation resources for empathetic dialogue \cite{xuan2025mmlu}, cross-linguistic pragmatic differences such as indirectness and \textit{aizuchi} in Japanese \cite{belay-etal-2025-culemo}, and language-specific segmentation and emotion semantics \cite{rusli2024experimental}. Thus, while current LLMs can generate appropriate validating responses, their affective calibration remains limited, particularly outside English.

To assess whether these automatic results reflect perceived therapeutic quality, we conducted a blinded human evaluation on 100 utterance--response pairs. As shown in Table~\ref{tab:human_eval_generation}, human reference responses received the highest scores on most dimensions, confirming that the benchmark targets are generally perceived as the strongest validating responses. Among the models, GPT is slightly stronger on \textit{Emotional Understanding} and \textit{Perceived Validation}, whereas Llama is slightly better on \textit{Non-judgmental Tone} and \textit{Mindful Presence}, though the differences are modest. The overall inter-annotator agreement, measured by Krippendorff's $\alpha$, is 0.394, reflecting the subjective nature of the task. These findings indicate that semantic similarity captures only part of validating quality. Case studies are provided in Appendix~\ref{sec:casestudy}.

% %
% \begin{table}[h]
% \centering
% \small
% \begin{tabular}{ll}
% \hline
% Utterance & Yeah they must practice a lot. I would be afraid of getting trampled \\

% \hline
% \end{tabular}
% \caption{Comparative case studies between different LLM baselines.}
% \label{tab:case}
% \end{table}
% %

% ### Compare the human evaluation between the best baseline and the ground truth ###

% \subsection{Discussion}

% \paragraph{Prompt strategy matters more than model size}
% Despite a 10× parameter gap to commercial GPT-4, the 8b Llama attains comparable BERT-F and higher Distinct-n when given three exemplars, emphasising the value of careful prompt design over sheer scale - especially in low-resource languages where training data are scarce.

% \paragraph{Diversity must be balanced against validation intent}
% High Distinct-n correlates with slightly lower IP/ER rates, echoing counselling guidelines that creativity should not compromise clear emotional acknowledgement. In practice, a hybrid prompt that first reasons (CoT) and then edits for brevity may capture both goals, an avenue we leave to future work.

% \paragraph{Cross-lingual generality remains challenging}
% Japanese BLEU is 5–7 pp higher than English, yet empathy-signal coverage lags in ER, hinting that literal translations of validating phrases (e.g., 「確かにね」) are under-represented in model pre-training corpora. Augmenting LLMs with culturally curated validation templates could close this gap.

\subsection{Joint Timing-and-Generation Evaluation}

EmoValidBench evaluates response generation assuming that validation is warranted, but a practical dialogue system must decide both \emph{when} to validate and \emph{what} to say, including for utterances that do not require validation. To assess this joint setting, we constructed a benchmark of 200 utterances (English and Japanese; per language, 50 validate and 50 non-validate cases). We compare four configurations: \textbf{MEGUMI+LLM}, which uses MEGUMI for timing and an LLM for generation; \textbf{LLM two-stage}, which uses an LLM for timing and another LLM for generation; \textbf{LLM multi-task}, where a single LLM first decides whether validation is needed and then generates a response only when appropriate; and \textbf{LLM alw. multi-task}, which uses the same format but always generates a validating response regardless of the timing decision. For each LLM-based setting, we use the same GPT 4.1 Nano backbone with 3-shots settings as in the response-generation experiments, with prompts provided in Appendix~\ref{sec:vrg}.

Table~\ref{tab:joint_timing_generation} shows a clear precision--recall trade-off in timing. \textbf{LLM alw. multi-task} achieves the highest timing recall (92.00) and the best timing F1 among the automatic systems (69.70), but does so by over-validating many negative cases. In contrast, \textbf{MEGUMI+LLM} is better calibrated for deciding \emph{when} to validate, achieving the highest timing precision (64.20) and the best message-level macro-F1 (61.15) among automatic systems, indicating fewer false positives on non-validation utterances. \textbf{LLM multi-task} also outperforms \textbf{LLM two-stage} on both message-level and timing-level F1, suggesting that joint modeling is more effective than a pipelined approach. Human judgments remain the upper bound (message-level F1: 70.82; timing-level F1: 75.78). Overall, these results support using a timing detector like MEGUMI to gate validation decisions, while relying on LLMs for response generation only when validation is warranted.

% \begin{table}[t]
% \centering
% \scriptsize
% \resizebox{\linewidth}{!}{%
% \begin{tabular}{l ccc c ccc}
% \toprule
% & \multicolumn{3}{c}{Message-level} & & \multicolumn{3}{c}{Timing-level} \\
% \cline{2-4}
% \cline{6-8}
% Model & Pre. & Rec. & F1 & & Pre. & Rec. & F1 \\
% \midrule
% Human Judgement & 74.50 & 71.67 & 70.82 && 66.20 & 88.67 & 75.78 \\
% MEGUMI + LLM & 61.93 & 61.50 & 61.15 && 64.20 & 52.00 & 57.46 \\
% LLM two-stage & 58.86 & 55.00 & 49.50 && 53.01 & 88.00 & 66.17 \\
% LLM multi-task & 62.06 & 58.00 & 54.14 && 55.06 & 87.00 & 67.44 \\
% LLM alw. multi-task & 66.94 & 60.00 & 55.44 && 56.20 & 92.00 & 69.70 \\
% \bottomrule
% \end{tabular}
% }
% \caption{Joint timing-and-generation evaluation result [\%]. Message-level scores evaluate the appropriateness of the final system behavior after timing and generation, while timing-level scores evaluate whether the system correctly decides whether validation is warranted.}
% \label{tab:joint_timing_generation}
% \end{table}
%on a balanced mixed benchmark containing both validate and non-validate utterances (200 utterances total; 100 EN, 100 JA). 

\begin{table}[t]
\centering
\scriptsize
\resizebox{\linewidth}{!}{%
\begin{tabular}{l cc c ccc}
\toprule
& \multicolumn{2}{c}{Message-level} & & \multicolumn{3}{c}{Timing-level} \\
\cline{2-3}
\cline{5-7}
Model & M-F1 & Bal. Acc. & & Pre. & Rec. & F1 \\
\midrule
Human Judgement       & 70.82 & 71.67 && 66.20 & 88.67 & 75.78 \\
MEGUMI + LLM          & 61.15 & 61.50 && 64.20 & 52.00 & 57.46 \\
LLM two-stage         & 49.50 & 55.00 && 53.01 & 88.00 & 66.17 \\
LLM multi-task        & 54.14 & 58.00 && 55.06 & 87.00 & 67.44 \\
LLM alw. multi-task   & 55.44 & 60.00 && 56.20 & 92.00 & 69.70 \\
\bottomrule
\end{tabular}
}
\caption{Joint timing-and-generation evaluation result [\%]. Message-level scores evaluate the appropriateness of the final system behavior after timing and generation, while timing-level scores evaluate whether the system correctly decides whether validation is warranted.}
\label{tab:joint_timing_generation}
\end{table}

\section{Conclusions}
In this work, we have presented the first comprehensive treatment of \textit{emotional validation} within dialogue systems, spanning task formalisation, data, models, and evaluation. We defined three clear subtasks - validating response identification, validation timing detection, and validating response generation - and introduced \textbf{M-EDESConv} and \textbf{M-TESC}, the first large-scale multilingual corpora annotated for validation phenomena in both text-based and spoken settings. Our proposed \textbf{MEGUMI} architecture leverages cross-lingual pretrained semantics together with language-specific emotion encoders, unified by cross-modal attention and a gated fusion mechanism, to accurately determine when a system should validate a user's feelings. We also introduced \textbf{EmoValidBench}, the first benchmark for validating response generation, providing evaluation scripts, LLM baselines, and empathy-oriented metrics to enable standardized comparisons across future models. Looking ahead, we would like to extend to the implementation of emotional validation in a conversational robot for the real-world study.

\section*{Limitations}

Despite these contributions, our study has several limitations. First, the scope of our curated data is confined to English and Japanese; other languages and cultural norms around emotional validation may exhibit different linguistic cues and pragmatic conventions that our current models cannot capture. Second, although we bootstrap annotation with a semi-automatic classifier to scale to 120 k turns, the reliance on confidence-filtered pseudo-labels carries the risk of residual errors and biasing downstream models, especially in low-resource or edge-case contexts. 
% Third, our validating response generation experiments rely exclusively on automatic metrics and empathy-signal classifiers; without human judgements of perceived empathy, naturalness, and user satisfaction, we cannot fully gauge the real-world effectiveness or potential unintended effects of generated replies.

Moreover, our timing-detection model operates solely on text transcripts and omits prosodic, acoustic, and visual cues known to inform validation in face-to-face interaction. The freeze of the XLM-RoBERTa backbone for computational tractability also precludes domain-specific fine-tuning that might further improve performance, and hardware constraints prevented exploration of larger language models beyond 8b parameters. Finally, while our experiments show promising performance in non-clinical dialogue, deploying emotional validation in sensitive domains such as mental-health support will require rigorous safety protocols, expert oversight, and continuous monitoring to avoid harm or overreliance on automated empathy.

\section*{Acknowledgments}
The authors would like to acknowledge Professor Mika Enomoto for providing us with access to the TUT Emotional Storytelling Corpus, which enabled us to analyze and draw conclusions from a vast amount of data. This work was also supported by JST Moonshot R\&D JPMJPS2011.
%\bibliography{anthology,custom}
% Custom bibliography entries only
\bibliography{custom}

\appendix

\section*{Appendix}

\section{Task Description}
\label{sec:task_description}

In this section, we will describe the task necessity for the emotional validation expression in the spoken dialogue system. Even though previous studies have shown that validation can be expressed through response generation~\cite{pang2024acknowledgment}, there aren't any formal task descriptions until now. Inspired by the theory of validation \cite{linehan1997validation}, we have defined the emotional validation in the spoken dialogue system into three subtasks, i.e. validating response identification, validation timing detection, and validating response generation. The summary of the task description we defined is summarized in the Figure~\ref{fig:framework}.

\subsection{Validating Response Identification}
The first requirement is to decide whether a system generated response is, validating, or known as \textbf{\textit{validating response}}. Mis-labeling brings risk: inappropriate reassurance or pseudo-empathy can increase user distress\footnote{\href{https://www.psychiatrictimes.com/view/when-validation-is-harmful}{https://www.psychiatrictimes.com/view/when-validation-is-harmful}} or alienation \cite{breslau1998trauma}. Linguistic studies of dialogue acts provide methodological precedents, showing that automatic classifiers can distinguish supportive acts such as ``appreciation'' or ``agreement'' \cite{welivita2020taxonomy, chen2022emphi} from neutral turns, but accuracy drops when acts overlap semantically \cite{stolcke2000dialogue, adiani2023dialogue}. Validation adds further nuance because the same surface pattern (e.g., ``I see'') may or may not affirm the user's emotion depending on context. Our corpus therefore begins with manual labels and expands them semi-automatically via a fine-tuned classifier, following hybrid annotation pipelines in emotion research \cite{canales2016innovative, fonteyn2024could}

\subsection{Validation Timing Detection}
Knowing when to validate is as critical as knowing how. Communication studies warn that over-frequent or ill-timed empathic moves can be perceived as insincere, reducing perceived provider empathy and therapeutic alliance \cite{roscoe2024timing, kuo2022and}. Similar timing effects emerge in social-robot experiments, where repetitive ``I understand'' statements without appropriate pauses diminished user rapport \cite{johanson2023effects}. Existing end-to-end generators seldom account for discourse-level timing; they optimise local next-utterance loss and may insert multiple empathic markers in rapid succession. We cast timing as a sequence-labeling task over the dialogue context, enabling models such as our MEGUMI architecture to decide whether the upcoming turn warrants validation.

\subsection{Validating Response Generation}
Finally, the system must produce a response that satisfies validation theory \cite{linehan1997validation}. Generic empathetic models often interleave advice, persuasion, or question-asking strategies that conflict with unconditional acknowledgment \cite{welivita2023empathetic, samad2022empathetic}. Moreover, validation can be expressed verbally and non-verbally; head nods, prosodic alignment, and empathic facial displays amplify perceived support in communications \cite{linehan1997validation, johanson2023effects, marcoux2024nonverbal}. Thus, we release the \textbf{EmoValidBench}, the first benchmark for validating response generation. It pairs each user turn that requires validation with evaluation scripts that measure semantic consistency, lexical diversity, and empathy-coverage.

\section{Related Work}

\subsection{Empathetic Response Generation}

Empathetic Response Generation (ERG)  began with EmpatheticDialogues (ED), which established emotion-grounded evaluation and showed that training on emotion-conditioned dialogues improves perceived empathy over chitchat \cite{rashkin2018towards}. Subsequent work emphasized affect matching, e.g., MoEL routes to listener experts by emotion, MIME mimics user affect, and knowledge/commonsense–augmented models (KEMP, CEM) inject situational priors to boost specificity \cite{lin2019moel, majumder2020mime, li2022knowledge, sabour2022cem}. Taxonomy-driven control labels empathetic intents to steer responses \cite{welivita2020taxonomy}.

Despite these advances, ERG only focus on generate empathetic response, which conflicts with emotional validation, that requires deciding whether/when to validate and delivering explicit acknowledgment rather than mirroring affect. The mismatch surfaces in support-oriented settings: ESConv frames multi-stage strategies (exploration, comforting, action), yet timing remains weakly supervised; recent analyses of LLMs on ESConv report strategy biases and difficulty choosing the right move, yielding shallow support \cite{liu2021towards, kang2024can}. Emerging work adds interpretable support structures (e.g., ESCoT, STEF) but still centers on style/strategy selection rather than principled validation decisions \cite{zhang2024escot, wang2023enhancing}.

We address the ERG–validation tension by separating decision from generation: first decide whether validation is warranted given context, then generate concise acknowledgment, yielding timing-aware, not purely affect-matching, support.

\begin{table*}[t]
\centering
\small
\begin{tabular}{lccccc}
\hline
Dataset & Modality & \#dialogue & \#utterance & Average \#word & Average \#turns \\
\hline
\rowcolor[gray]{0.95}\textbf{EmpatheticDialogues} &  &  &  & & \\
\textit{-English} & Text & 24.8k & 82k & 15.2 & 4.31  \\
\textit{-Japanese}  & Text & 20k & 80k & 25 & 2 \\
\rowcolor[gray]{0.95}\textbf{ESConv} &  &  &  & & \\
\textit{-English} & Text & 1.3 k & 17.6k & 15.19 & 13.62 \\
\textit{-Japanese}  & Text & 1.3 k & 17.6k & 21.84 & 13.62 \\
\rowcolor[gray]{0.95}\textbf{TESC} &  &  &  & & \\
\textit{-English}& Speech & 247 & 3080 & 34.85 & 8 \\
\textit{-Japanese} & Speech &247 & 3080 & 41 & 8 \\
\hline
\end{tabular}
\caption{Statistics of the English and Japanese splits of the three base corpora employed in this study.}
\label{tab:ori_data}
\end{table*}

\subsection{Emotional Validation}

Clinical psychology defines emotional validation as communicating that a person's feelings ``make sense'' in context. Dialectical Behavior Therapy (DBT) details graded validation levels, underscoring timing and contextual fit as core to therapeutic alliance, distinct from generic empathy \cite{linehan1997validation}. Computational treatment is nascent. Prior work combining emotion recognition and phrase-based generation shows feasibility on Japanese ED (text) and the TUT Emotional Storytelling Corpus (speech), with task-adaptive BERT outperforming strong baselines \cite{pang2024acknowledgment}. Complementary analyses on TESC find acoustic/linguistic cues predictive of validation-worthy moments, reinforcing the primacy of timing \cite{pang2023prediction}.

Despite these demonstrations, prior work lacks a standardized task description and benchmark for emotional validation in spoken dialogue systems, leaving the problem under-specified relative to ERG datasets. Inspired by validation theory \cite{linehan1997validation}, we formalize emotional validation for spoken dialogue as three subtasks: validating response identification, validation timing detection, and validating response generation, providing a clinically grounded decomposition compatible with ERG pipelines.

% %
% \begin{table*}[h]
% \centering
% \small
% \begin{tabular}{lccccccc}
% \hline
% \multicolumn{1}{c}{} & \multicolumn{3}{c}{Macro Average} & & \multicolumn{3}{c}{Target Class} \\
% \cline{2-4}
% \cline{6-8}
% \multicolumn{1}{c}{} & Precision & Recall & F1-Score & & Precision & Recall & F1-Score \\ \hline
% Random Baseline & 50.00 & 50.00 & 48.38 & & 32.45 & 50.13 & 39.40 \\
% mBERT & 74.33 & 74.51 & 74.30 & & 70.24 & 76.32 & 73.15\\
% Llama 3.1 8b &  &  &  & &  &  &  \\
% \textit{- Zero-shot} & 54.85 & 53.39 & 41.04 & & 34.37 & 85.61 & 49.04 \\
% \textit{- 3-shot} & 61.03 & 52.77 & 32.18 & & 33.82 & \textbf{97.88} & 50.27 \\
% \textit{- LoRA} & 58.36 & 51.81 & 32.96 & & 37.16 & 97.02 & 53.74\\
% GPT 4.1 Nano &  &  &  & &  &  &  \\
% \textit{- Zero-shot} & 64.49 & 65.04 & 57.98 & & 42.71 & 85.19 & 56.89\\
% \textit{- 3-shot} & 67.21 & 69.57 & 66.57 & & 50.36 & 74.60 & 60.13\\
% % XLM-RoBERTa & 81.36 & 81.12 & 80.37 & & 73.31 & 89.85 & 80.74\\
% XLM-RoBERTa & \textbf{86.42} & \textbf{85.30} & \textbf{85.28} & & \textbf{80.66} & 93.64 & \textbf{86.67}\\

% \hline
% \end{tabular}
% \caption{Results of validating response identification task (automatic annotation) [\%].}
% \label{tab:automatic_annotation}
% \end{table*}
% %

\section{Base Corpora}
\label{sec:appendix}

EmpatheticDialogues (ED) contains 24.8k two-speaker conversations collected from crowdworkers imagining specific emotional situations \cite{rashkin2018towards}. ESConv complements ED with 1,053 counselor-style dialogues in which trained volunteers comfort users facing real-life stressors \cite{liu2021towards}. To further evaluate under the spoken dialogue scenario, we include the TUT Emotional Storytelling Corpus (TESC)~\cite{oishi2021design}, a Japanese two-party, multi-turn corpus (247 sessions; $\approx$9.2 h) where close friends recount experiences under eight Plutchik emotion prompts~\cite{plutchik2001nature}. 

Table \ref{tab:ori_data} summarises the size, modality and interactional profile of the three corpora that constitute our base dataset. EmpatheticDialogues (ED) offers the broadest coverage with 24.8k English and 20k Japanese text conversations; yet these crowdsourced exchanges are succinct - just 4.3 and 2.0 turns on average, with roughly 15–25 tokens per utterance - because speakers were asked to recount short personal stories to a listener who responds empathetically.
ESConv contributes 1.3 k expert-annotated emotional-support dialogues per language. Although an order of magnitude smaller than ED, each session resembles real counselling, spanning ≈ 14 turns; Japanese utterances are notably longer (≈ 22 tokens) than English (≈ 15), matching prior observations on script complexity in bilingual corpora.
Finally, the TUT Emotional Storytelling Corpus (TESC) introduces the spoken modality with 247 transcribed sessions per language. The oral setting yields markedly longer utterances (≈ 35 tokens in English, 41 in Japanese) while keeping the turn budget concise at eight per dialogue.

\section{Translation Quality Evaluation}
\label{sec:quality}

As part of our data construction method involved the automatic translation from the LLMs, it is neccessary to evaluate the quality of these synthetic data.

\subsection{Objective Quality Evaluation}
We first conducted an objective evaluation. Since we do not have a dataset that consist of reference that allow us to use the normal machine traslation metric like BLEU score, we decided to evaluate through the COMETKiwi metric \cite{rei2022cometkiwi}. COMETKiwi is a reference-free model employs a regression approach and is built on top of InfoXLM. It has been trained using direct assessments from WMT17 to WMT20, as well as direct assessments from the MLQE-PE corpus. It provides scores ranging from 0 to 1, where 1 signifies a perfect translation. We have evaluated on both utterance and response perspective, i.e. evaluate the translation quality for the generated utterance and generated response. As a result, The translations scored 0.8470 for utterances and 0.8479 for responses, indicating high overall quality for dataset construction.

\subsection{Subjective Quality Evaluation}
We also conducted another human assessment on top of our objective evaluation. We recruited three bilingual Japanese--English speakers through crowdsourcing and asked them to evaluate sampled translations using five criteria: \textit{Accuracy}, which measures how closely the translation preserves the meaning of the source text; \textit{Contextual Consistency}, which evaluates consistency of terminology and style across the dialogue; \textit{Emotional Consistency}, which assesses whether the intended emotion and tone are preserved; \textit{Fluency and Readability}, which measures how natural and easy the translation is for a native speaker of the target language; and \textit{Cultural Appropriateness}, which evaluates whether the translation remains culturally suitable for the target audience. To prevent the biasness by the annotaters, we also calculated the inter-annotator agreement (IAA). The average scores are shown in Table~\ref{tab:translation_human_eval}. Across all dimensions, the translations received a mean score of 5.63, with the highest score on Fluency and Readability (5.72) and consistently strong ratings on emotion-related dimensions such as Emotional Consistency (5.61) and Cultural Appropriateness (5.64). These results complement the COMETKiwi scores and indicate that the translation workflow preserves not only semantic content, but also affective tone and cultural appropriateness.

\begin{table}[h]
\centering
\small
\begin{tabular}{lcc}
\toprule
Criterion & Average & IAA \\
\midrule
Accuracy & 5.65 & 0.392 \\
Contextual Consistency & 5.52 & 0.407 \\
Emotional Consistency & 5.61 & 0.263 \\
Fluency and Readability & 5.72 & 0.318 \\
Cultural Appropriateness & 5.64 & 0.339 \\
\midrule
Overall & 5.63 & 0.344 \\
\bottomrule
\end{tabular}
\caption{Human evaluation results for translation quality. Three bilingual Japanese--English annotators rated the translated data on a Likert scale across five dimensions.}
\label{tab:translation_human_eval}
\end{table}

\section{Validating Response Identification}
\label{sec:identification}

\subsection{Training Hyperparameters}
\label{sec:identify_para}

We fine-tune the \textit{xlm-roberta-large} model \cite{Alexis2019} with a learning rate of $1\times10^{-5}$, a batch size of 64, and train for 20 epochs. Evaluation is performed every 200 steps, using the Adam optimizer with L2 regularization (weight decay of 0.01). Early stopping is applied with a patience of 5 epochs. To ensure high precision for the validation class, we apply a confidence threshold of 0.75 during inference. 

\subsection{Comparative Baselines}
As part of our ablation study, we compare the fine-tuned \textit{xlm-roberta-large} model against several baselines, including a random baseline, multilingual BERT (mBERT)\footnote{\href{https://huggingface.co/google-bert/bert-base-multilingual-cased}{https://huggingface.co/google-bert/bert-base-multilingual-cased}}, LLaMA 3.1 8B\footnote{\href{https://huggingface.co/meta-llama/Llama-3.1-8B-Instruct}{https://huggingface.co/meta-llama/Llama-3.1-8B-Instruct}}, and GPT-4.1-nano\footnote{\href{https://openai.com/index/gpt-4-1/}{https://openai.com/index/gpt-4-1/}}. 

\begin{table*}[h]
\centering
\small
\begin{tabular}{lcccc c cccc}
\hline
                & \multicolumn{4}{c}{M-EDESConv} & & \multicolumn{4}{c}{Human-validated subset} \\ \cline{2-5} \cline{7-10}
                & \multicolumn{1}{c}{Overall} & \multicolumn{3}{c}{Target Class} & & \multicolumn{1}{c}{Overall} & \multicolumn{3}{c}{Target Class} \\ \cline{2-5} \cline{7-10}
                & M-F1 & Precision & Recall & F1-Score & & M-F1 & Precision & Recall & F1-Score \\ \hline

Random Baseline      & 48.38 & 32.45 & 50.13 & 39.40 & & 51.23      &  51.25     &  51.75     &  51.48       \\
mBERT                & 74.30 & 70.24 & 76.32 & 73.15 & & 72.96 & 71.30 & 77.00 & 74.04 \\
Llama 3.1 8b         &       &       &       &       & & &       &       & \\
\textit{- Zero-shot} & 41.04 & 34.37 & 85.61 & 49.04 & & 52.13 & 54.14 & 87.60 & 66.92       \\
\textit{- 3-shot}    & 32.18 & 33.82 & \textbf{97.88} & 50.27 & & 46.58 & 53.26 & 98.00 & 69.01              \\
\textit{- LoRA}      & 32.96 & 37.16 & 97.02 & 53.74 & &  44.05     &  51.67     & 93.00      &  66.43     \\
GPT 4.1 Nano         &       &       &       &       & & &       &       & \\
\textit{- Zero-shot} & 57.98 & 42.71 & 85.19 & 56.89 & & 72.06     &  73.33     &  69.60     & 71.36      \\
\textit{- 3-shot}    & 66.57 & 50.36 & 74.60 & 60.13 & & 61.90 & 59.57 & 84.00 & 69.71             \\
XLM-RoBERTa          & \textbf{85.28} & \textbf{80.66} & 93.64 & \textbf{86.67} & & 82.43 & 78.76 & 89.00 & 83.57 \\
Human                &   \multicolumn{4}{c}{No Applicable}      & & 85.10 & 80.94 & 92.00 & 86.12 \\ \hline

\end{tabular}
\caption{Results of validating response identification task [\%]. The left block reports results on the M-EDESConv dataset for automatic annotation, and the right block reports performance on the human-validated subset.}
\label{tab:automatic_annotation}
\end{table*}

\subsection{Result and Discussion}
\label{sec:identify_result}

Table \ref{tab:automatic_annotation} reports the performance of several automatic classifiers that were used to propagate validating response labels from a manually annotated seed set to the full M-EDESConv corpus. All scores are averaged over five random initialisations; we present both macro-averaged metrics (capturing overall label balance) and scores for the minority validate class, which is the critical signal for downstream tasks.

The fine-tuned XLM-RoBERTa model clearly emerges as the most reliable annotator. It achieves 85.3\% macro F1 and 86.7\% target-class F1, outperforming the next-best baseline (mBERT) by roughly eleven points on each metric. Precision gains (+10.4 pp over mBERT) indicate fewer false positives, while the high recall of 93.6\% demonstrates that the model rarely misses genuine validating utterances - essential for minimising label noise.
Instruction-tuned LLMs require careful prompting to approach Pre-trained Language Model (PLM) performance. In zero-shot mode, GPT-4.1 nano delivers respectable macro F1 (58.0\%) but suffers from low precision (42.7\%) on the validation class, leading to many spurious positives. Providing three in-context examples narrows the gap (macro F1 = 66.6\%), yet precision (50.4\%) still trails far behind XLM-RoBERTa. Llama-3.1 8B exhibits a complementary error profile: three-shot prompting attains the highest recall in the table (97.9\%) but collapses precision to 33.8\%, effectively labelling almost every response as validating and therefore offering little discriminative value.

These results motivated our choice of the XLM-RoBERTa classifier for corpus-wide pseudo-labelling. To mitigate residual noise we retained only predictions with confidence ≥ 0.90, yielding a class distribution that closely mirrors the manually annotated subset (described in \ref{sec:3}). Although LLMs currently lag behind supervised PLMs for this task, their high recall could still prove useful in an ensemble or active-learning setting - an avenue we leave for future work.

\section{Human Evaluation on Automatic Annotation}
\label{sec:human_eval_auto}
As 94.11\% of M-EDESConv was labeled automatically, we conducted an additional human evaluation study to quantify the reliability of the pseudo-labels. For each language (English and Japanese), we randomly sampled 100 automatically annotated responses, balanced across classes (50 validate and 50 non-validate), and collected three native-speaker judgments via crowdsourcing on whether each response was validating. The human judgments agreed with the automatic labels in 85.0\% of cases on average. Agreement between individual annotators and the automatic labels ranged from 83--88\% for English and 84--87\% for Japanese. We also measured inter-annotator agreement among the human raters. The average pairwise Cohen's $\kappa$ was 0.752 overall (English: 0.740/0.744/0.706; Japanese: 0.819/0.820/0.681), indicating substantial agreement. 
% These results suggest that the automatic labels are sufficiently reliable for large-scale corpus construction, while still leaving some ambiguity in difficult cases and motivating our continued use of manually annotated evaluation sets.

Table~\ref{tab:automatic_annotation} further compares model performance on the manually verified sample. Among the automatic methods, fine-tuned XLM-RoBERTa performs best, achieving 82.43 macro-F1 and 83.57 validation-class F1, outperforming mBERT and the compact LLM baselines. Its performance is also closest to human performance (85.10 macro-F1; 86.12 validation-class F1), and thus, shown the effectiveness of choosing XLM-RoBERTa as the backbone for the automatic annotation in this study.

% %
% \begin{table}[t]
% \centering
% \small
% \begin{tabular}{l|ccc|ccc}
% \toprule
% & \multicolumn{3}{c|}{Macro} & \multicolumn{3}{c}{Validation class} \\
% Model & P & R & F1 & P & R & F1 \\
% \midrule
% mBERT & 73.15 & 73.00 & 72.96 & 71.30 & 77.00 & 74.04 \\
% XLM-RoBERTa & 83.06 & 82.50 & 82.43 & 78.76 & 89.00 & 83.57 \\
% Llama 3.1 8B & 70.38 & 56.00 & 46.58 & 53.26 & 98.00 & 69.01 \\
% GPT-4.1 nano & 66.23 & 63.50 & 61.90 & 59.57 & 84.00 & 69.71 \\
% Human & 85.84 & 85.17 & 85.10 & 80.94 & 92.00 & 86.12 \\
% \bottomrule
% \end{tabular}
% \caption{Performance on the manually verified sample used to assess annotation quality. Macro denotes macro-average over the validate/non-validate classes; Validation class reports scores for the positive (validate) class.}
% \label{tab:human_verification_annotation_quality}
% \end{table}
% %

\section{Annotated Dataset Details}
\label{sec:dataset_details}

To preserve label distribution consistency with the manually annotated subset, we analyze the prediction confidence scores across each sub-dataset. Based on this analysis, we set confidence thresholds of 0.90 for ED and 0.95 for ESConv to align the automated annotations with the original distribution. As a result, the total dataset comprises 98.6k responses, including 81k from EmpatheticDialogues and 17.6k from ESConv. Among these, 5,805 responses were manually annotated, where 1,204 validating and 1,663 non-validating (total: 2,867) from EmpatheticDialogues, and 680 validating and 2,258 non-validating (total: 2,938) from ESConv. This means manual annotations account for 5.89\% (5,805) of the dataset, while the remaining 94.11\% (92,795 responses) were annotated automatically using our trained validation classifier. 

\hspace*{0pt} 
\begin{table}[htbp!]
\scalebox{0.75}{
\begin{tabular}{lll|rrr}
\hline
Languages                 & Models                              & Traits & Acc.  & BA.   & F1    \\ \hline
\multirow{3}{*}{English}  & \multirow{3}{*}{RoBERTa}       & ER     & 84.76 & 84.13 & 84.70 \\
                          &                                     & IP     & 84.12 & 85.35 & 84.23 \\
                          &                                     & EX     & 94.81 & 92.46 & 94.86 \\ \hline
\multirow{3}{*}{Japanese} & \multirow{3}{*}{LUKE Japanese} & ER     & 73.74 & 72.64 & 73.52 \\
                          &                                     & IP     & 79.09 & 79.29 & 79.22 \\
                          &                                     & EX     & 88.82 & 77.37 & 88.27 \\ \hline
\multirow{3}{*}{Both}     & \multirow{3}{*}{XLM-RoBERTa}   & ER     & 77.88 & 76.66 & 77.61 \\
                          &                                     & IP     & 81.28 & 82.13 & 81.42 \\
                          &                                     & EX     & 91.82      &  85.08     & 91.69      \\ \hline
\end{tabular}}
\caption{Evaluation results of the empathetic signal predictors. Acc., BA., and F1 refer to accuracy, balanced accuracy, and weighted F1 score, respectively.}
\label{tab:empathyaccu}
\end{table}

\section{Experimental Settings in Validation Timing Detection}
\label{sec:setting_vtd}

We fine-tuned all models on the M-EDESConv corpus with a learning rate of $1\times10^{-5}$, a batch size of 64 (with gradient accumulation over 8 steps), and a 20-epoch cap. To regularise training we combined L2 normalization (weight decay rate of 0.01) with early stopping after five stagnant validation checks. Validation was run every 250 steps and the best checkpoint was selected by F1. Five random seeds were used throughout to mitigate variance.

\section{Empathetic Signal Coverage Training Details}
\label{sec:empathytrain}

For English, we follow the official annotation schema\footnote{\href{https://github.com/behavioral-data/Empathy-Mental-Health}{https://github.com/behavioral-data/Empathy-Mental-Health}} and fine-tune three RoBERTa models \cite{liu2019roberta}, each followed by a linear classifier, to determine whether a response conveys each signal. For Japanese, we translate the English corpus by GPT-4.1-mini with temperature set to 0.1, and then classify the signals using a LUKE model\footnote{\href{https://huggingface.co/studio-ousia/luke-japanese-base}{https://huggingface.co/studio-ousia/luke-japanese-base}} followed by two linear layers. In the multilingual setting, we employ XLM-RoBERTa \cite{Alexis2019} with two linear layers as a unified classifier across languages.

We set the batch size to 32, the learning rate to 2 × 10$^{-5}$, and the dropout rate in the linear layers to 0.1, and train for up to seven epochs with early stopping after three epochs without improvement. The evaluation results of the empathetic-signal predictors are summarized in Table~\ref{tab:empathyaccu}.

\section{Case Studies}
\label{sec:casestudy}

The case studies in Table~\ref{tab:case} illustrate the same pattern discussed in Section~\ref{sec:response_result}. Model outputs are often close to the gold responses in meaning, as reflected by high BERTScore, even when surface forms differ. For example, for ``Yeah they must practice a lot. I would be afraid of getting trampled,'' Llama~3.1~8b's ``I can understand that. I would feel the same way'' and GPT-4's ``That makes sense; it's understandable to feel worried about that'' both preserve the validating intent of the gold response, ``Me too, that would be terrible.''

\begin{table}[t]
  \centering
  \small
  \begin{tabularx}{0.5\textwidth}{lX}
  \hline
    \textbf{Model} & \textbf{Text} \\
    \hline
    Utterance & Yeah they must practice a lot. I would be afraid of getting trampled \\
    Ground Truth & Me too, that would be terrible. \\
    \hline
    Llama 3.1 8B & I can understand that. I would feel the same way. \\
    GPT-4.1 Nano & That makes sense, it's understandable to feel worried about that. \\
    \hline
  \end{tabularx}
  \caption{Comparative case studies between different LLM baselines.}
\label{tab:case}
\end{table}

\section{Prompts}
\label{sec:prompts}
\subsection{English-Japanese translation}
\label{sec:en-ja}
You are a professional Japanese translator.
For the following English utterance, please translate into natural, spoken‐style daily Japanese as a native speaker would say.
You should avoid literal word‐for‐word renderings.

\subsection{Japanese-English translation}
\label{sec:ja-en}
You are a professional English translator.
For the following Japanese utterance, please translate into natural, spoken‐style daily English as a native speaker would say.
You should avoid literal word‐for‐word renderings.

\subsection{Validating Response Identification}
\label{sec:vti}

Definition of validation: Validation is a communication technique, where we recognize, understand, and acknowledge others' emotional states, thoughts, and actions. \\

\noindent Please classify the response into validating or non-validating response. Return validate if it is a validating response and non-validate if it is a non-validating response.\\

\noindent Followed by the three examples dialogues with validating response, and another three examples dialogues with non-validating response in each language, respectively.

\subsection{Validation Timing Detection}
\label{sec:vtd}
Definition of validation: Validation is a communication technique, where we recognize, understand, and acknowledge others' emotional states, thoughts, and actions. \\

\noindent Please classify each utterance into whether a validating response should be generated. Return validate if needed to generate a validating response and non-validate if not necessary to generate (meaning that it will generate a non-validating response)\\

\noindent Followed by the three examples dialogues with validating response, and another three examples dialogues with non-validating response in each language, respectively. Let's think step by step.

\subsection{Validating Response Generation}
\label{sec:vrg}
Definition of validation: Validation is a communication technique, where we recognize, understand, and acknowledge others' emotional states, thoughts, and actions. \\

\noindent You should act as a listener, in speech conversations. Please generate a validating response for the given utterances from the speaker.
The generated response should be a validating response. You should only respond with a validating response, excluding other information (without Listener:).\\

\noindent Followed by three examples dialogues with validating response in each language, respectively. Let's think step by step.

\subsection{LLM-as-judge Evaluation}
\label{sec:llmjudgeprompt}
% \begin{quote}
\textbf{System:} You are an expert evaluator of \textsc{emotional validation} in supportive conversations. Score the response for how well it validates the user's feelings. Be concise, neutral, and avoid offering new counseling. Use the NURSE and OARS principles. Return a structured output only.

\textbf{User:} Evaluate how well the \texttt{RESPONSE} validates the user's emotions.

\texttt{USER:} \{user utterance\}

\texttt{RESPONSE:} \{candidate response\}

\textbf{Rubric:}
\begin{itemize}
    \item \texttt{acknowledges\_feelings}: names or mirrors the user's feelings
    \item \texttt{accurate\_reflection}: reflects the user's emotion and context without distortion
    \item \texttt{respect\_nonjudgment}: uses respectful, non-judgmental language
    \item \texttt{support\_warmth}: conveys support, compassion, and validation
    \item \texttt{safety\_boundaries}: avoids unsafe advice, overclaiming, or inappropriate boundary violations
\end{itemize}

Return per-criterion scores, short explanations, a weighted overall score, and a final verdict.

\end{document}